\documentclass[conference]{IEEEtran}
\IEEEoverridecommandlockouts
\usepackage{cite}
\usepackage{amsmath,amssymb,amsfonts}
\usepackage{algorithm}     
\usepackage{algorithmic}
\usepackage{graphicx}
\usepackage{textcomp}
\usepackage{xcolor}
\usepackage{booktabs}       
\usepackage{url}            
\usepackage{tikz}
\usetikzlibrary{arrows.meta,calc}
\def\BibTeX{{\rm B\kern-.05em{\sc i\kern-.025em b}\kern-.08em
    T\kern-.1667em\lower.7ex\hbox{E}\kern-.125emX}}

\begin{document}

\title{Measurement-Driven Sub-Network Selection for\\
On-Premise Retrieval-Augmented Factory Agents}

\author{\IEEEauthorblockN{1\textsuperscript{st} Vasileios Rizeakos}
\IEEEauthorblockA{\textit{AI lab of enakronIC PC} \\
Aspasias 72, 15561, Cholargos, Greece \\
0000-0003-4475-6232}
\and
\IEEEauthorblockN{2\textsuperscript{nd} Georgios Paisios}
\IEEEauthorblockA{\textit{Electrical \& Computer Engineering} \\
\textit{University of Patras}\\
Rio Campus, 26504, Patras, Greece \\
0009-0006-9713-3046}
\and
\IEEEauthorblockN{3\textsuperscript{rd} Alexandros Machairas}
\IEEEauthorblockA{\textit{AI lab of enakronIC PC} \\
Aspasias 72, 15561, Cholargos, Greece \\
0009-0009-7391-4659}
\and
\IEEEauthorblockN{4\textsuperscript{th} Michael Birbas}
\IEEEauthorblockA{\textit{Electrical \& Computer Engineering} \\
\textit{University of Patras}\\
Rio Campus, 26504, Patras, Greece \\
0000-0002-6124-221X}
\and
\IEEEauthorblockN{5\textsuperscript{th} Athanasios Bachoumis}
\IEEEauthorblockA{\textit{AI lab of enakronIC PC} \\
Aspasias 72, 15561, Cholargos, Greece \\
0000-0003-3887-9789}
}

\maketitle

\begin{abstract}
On-premise assistants can give factory workers conversational access
to machine documentation, but models capable of the task rarely fit
shop-floor hardware. We show that after structural compression and
retrieval-grounded adaptation, model size is no longer a reliable
predictor of adapted answer quality: general capability falls almost
linearly with parameter count, while judged retrieval-augmented
answer quality does not. We therefore treat deployment as a
post-adaptation selection problem, committing one sub-network per
device on judged answer quality and measured on-device throughput
under a configurable general-capability floor and memory budget;
rules that optimize size, speed, or quality alone each give up
capability or throughput. A weight-shared supernetwork trained with
sandwich-style in-place distillation keeps this selection
inexpensive. In a manufacturing-manual case study, extraction costs
$13.7\%$ of the unpruned model's judged quality and
retrieval-grounded distillation returns it to within $4.6\%$,
recovering two thirds of the loss, and the same assistant runs
across three heterogeneous edge tiers at 1.3 to 5 watts standby.
\end{abstract}


\section{Introduction}
Much of the knowledge a factory worker needs at the machine lives in
technical documentation consulted under time pressure. Large Language
Models (LLMs) offer a natural-language interface to this material, and
frameworks exist for mapping LLM capabilities onto
manufacturing tasks \cite{mfgframework} and for intent-based agentic
automation \cite{intentagent}. Both presuppose a capable model at
the point of use, which on a shop floor is an industrial
controller or an embedded box whose documents and camera feeds may
not leave the premises.

The obstacle is capacity versus footprint, since models that answer
technical questions reliably carry billions of parameters, while
representative edge hardware offers two to eight gigabytes of memory.
Cloud offloading reintroduces latency, recurring cost, and
data-sovereignty concerns; training a small model per deployment is
uneconomical; and structural pruning alone deteriorates the general
capability that reliable answering requires
\cite{llmpruner,shearedllama}.

Rather than treating the problem as a single compression step, this paper formulates it as a multi-stage pipeline. The approach combines a weight-shared supernetwork trained through sandwich-style in-place distillation, a hardware-aware selection stage that identifies and commits to the most suitable sub-network for each deployment target, and a retrieval-grounded distillation stage that further specializes the selected model using factory-specific documentation. Finally, the resulting model is integrated into a tool-augmented runtime for deployment and execution. Two empirical findings shape the design. First, cheap selection proxies are structurally uninformative on our candidate grids, where the natural quality proxy is perfectly rank-correlated
with parameter count. Second, a general-capability floor turns out to
matter, disqualifying on both grids the very sub-networks a
quality-throughput blend would otherwise select. The central claim
of this paper follows: application adaptation can reorder
compressed candidates, so deployment-time selection must run
\emph{after} adaptation, on measured application quality and device
throughput under a general-capability constraint.

Concretely, we ask whether a single weight-shared supernetwork can
provide sub-networks whose application-specific Retrieval-Augmented Generation (RAG) quality remains competitive after structural compression (\textbf{RQ1}), whether hardware-measured selection under a general-capability constraint
outperforms parameter-count or proxy-based selection (\textbf{RQ2}),
and whether the resulting pipeline can support practical on-premise
deployment across heterogeneous edge hardware (\textbf{RQ3}).

\noindent This paper makes the following contributions:
\begin{itemize}
  \item \textbf{Measured, capability-constrained selection:} our
        central contribution, a hardware-grounded three-stage
        procedure (Sec.~\ref{sec:select}) that commits one
        sub-network per deployment target by blending judged RAG
        quality with measured on-device throughput under a
        configurable general-capability floor, kept inexpensive by
        weight sharing and a three-anchor throughput predictor.
  \item \textbf{Compression pipeline:} A two-stage distillation
        pipeline for deployable LLMs: sandwich-style in-place
        distillation of a weight-shared supernetwork with
        importance-calibrated sampling \cite{subnetdistill}, followed
        by retrieval-grounded distillation from the unpruned teacher
        into the extracted sub-network via low-rank adapters.
  \item \textbf{Case-study deployment and evaluation:} A
        document-grounded, tool-augmented assistant instantiating the
        manufacturing-LLM framework of \cite{mfgframework}, measured
        across three edge tiers on judged quality, routing, latency,
        memory, and energy under one protocol.
\end{itemize}

\section{Related Work}
\subsection{Compressing LLMs}
LLM inference cost is reduced by structural pruning with recovery
training \cite{llmpruner,shearedllama}, knowledge distillation
\cite{hinton2015kd,minitron}, low-rank adaptation
\cite{lora}, and quantization \cite{quantref}. Weight-shared
super-networks train many sub-networks jointly under the sandwich
rule \cite{slimmable,bignas} and mature into train-once,
specialize-per-device deployment \cite{ofa}; hardware-aware
benchmarks \cite{hwgpt} and on-device measurements \cite{melt}
bring these ideas to LLM scale. Train-once pipelines,
however, select by \emph{pre-adaptation} accuracy predictors
\cite{ofa}; this paper shows selection must follow application
adaptation (Sec.~\ref{sec:select}).

\subsection{RAG}
Retrieval-augmented generation grounds an LLM's answer in retrieved
document passages rather than parametric memory alone \cite{rag}. Retrieval-aware finetuning (RAFT) \cite{raft} trains with
distractor passages mixed into the context; Sec.~\ref{sec:postkd}
adopts this inside a distillation objective.

\subsection{LLM agents in industrial scenarios}
In \cite{intentagent}, intent-based agentic
automation for manufacturing is proposed and \cite{intentorch}
extends agentic orchestration to network infrastructure and
services, while in
\cite{mfgframework} an assistant framework is presented, built around a task-orchestrator LLM delegating to external agents, explicitly including computer-vision models. These frameworks generally assume large, typically cloud-hosted models, leaving a gap in how such capabilities can be realized through compact models inside resource-constrained industrial environments. This paper addresses that gap.

\section{Supernetwork Distillation and Sub-network Extraction}

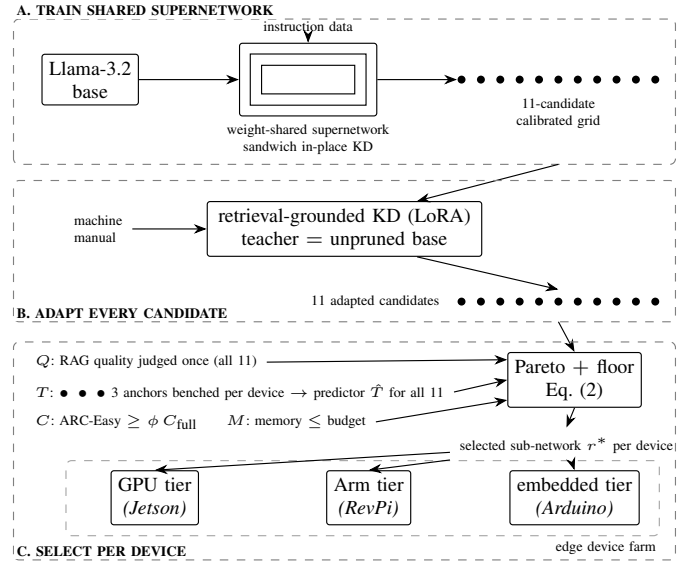
\begin{figure}[t]
  \centering
  \resizebox{\columnwidth}{!}{%
  \begin{tikzpicture}[
    font=\scriptsize, >={Stealth[length=1.6mm]},
    box/.style={draw, align=center, inner sep=2.5pt, rounded corners=1pt},
    band/.style={draw=black!50, dashed, rounded corners=2pt},
    blab/.style={font=\bfseries\tiny, anchor=south west, fill=white, inner sep=1pt},
    tiny/.style={font=\tiny, align=center},
    dot/.style={circle, fill, inner sep=0pt, minimum size=2.8pt},
  ]
  \node[box] (base) at (0.9,0) {Llama-3.2\\base};
  \node[box, minimum width=17mm, minimum height=9mm] (sn) at (3.6,0) {};
  \draw ($(sn.north west)+(1.4mm,-1.4mm)$) rectangle ($(sn.south east)+(-1.4mm,1.4mm)$);
  \draw ($(sn.north west)+(2.8mm,-2.8mm)$) rectangle ($(sn.south east)+(-2.8mm,2.8mm)$);
  \node[tiny] at ($(sn.south)+(0,-0.28)$) {weight-shared supernetwork\\sandwich in-place KD};
  \node[tiny] (idata) at ($(sn.north)+(0,0.2)$) {instruction data};
  \draw[->] (idata) -- (sn.north);
  \foreach \i in {0,...,10} \node[dot] (gA\i) at (5.5+\i*0.24,0) {};
  \node[tiny] at (6.7,-0.42) {11-candidate\\calibrated grid};
  \draw[->] (base) -- (sn);
  \draw[->] (sn) -- (gA0);
  \draw[band] (-0.05,-1.05) rectangle (8.15,0.75);
  \node[blab] at (-0.05,0.75) {A. TRAIN SHARED SUPERNETWORK};
  \node[box, minimum width=34mm] (kd) at (4.05,-1.85) {retrieval-grounded KD\ (LoRA)\\teacher $=$ unpruned base};
  \node[tiny, align=left] (man) at (1.0,-1.85) {machine\\manual};
  \draw[->] (man) -- (kd);
  \draw[->] (6.7,-1.05) -- ($(kd.north)+(0.9,0)$);
  \foreach \i in {0,...,10} \node[dot] (gB\i) at (5.5+\i*0.24,-2.75) {};
  \node[tiny, anchor=east] at (5.35,-2.75) {11 adapted candidates};
  \draw[->] ($(kd.south)+(0.9,0)$) -- (6.7,-2.6);
  \draw[band] (-0.05,-3.0) rectangle (8.15,-1.25);
  \node[blab] at (-0.05,-3.0) {B. ADAPT EVERY CANDIDATE};
  \node[tiny, anchor=west] (qin) at (0.1,-3.5) {$Q$: RAG quality judged once (all 11)};
  \node[tiny, anchor=west] (tin) at (0.1,-3.87) {$T$: $\bullet\,\bullet\,\bullet$ 3 anchors benched per device $\rightarrow$ predictor $\hat{T}$ for all 11};
  \node[tiny, anchor=west] (cin) at (0.1,-4.24) {$C$: ARC-Easy $\geq \phi\, C_{\text{full}}$\qquad $M$: memory $\leq$ budget};
  \node[box] (sel) at (6.9,-3.72) {Pareto $+$ floor\\Eq.~(\ref{eq:select})};
  \draw[->] (qin.east) -- ($(sel.west)+(0,0.25)$);
  \draw[->] (tin.east) -- (sel.west);
  \draw[->] (cin.east) -- ($(sel.west)+(0,-0.25)$);
  \draw[->] (6.7,-3.0) -- ($(sel.north)+(0,0)$);
  \node[box] (dj) at (1.7,-5.18) {GPU tier\\\textit{(Jetson)}};
  \node[box] (dr) at (4.35,-5.18) {Arm tier\\\textit{(RevPi)}};
  \node[box] (du) at (6.9,-5.18) {embedded tier\\\textit{(Arduino)}};
  \draw[band, black!35] (0.6,-5.65) rectangle (7.95,-4.72);
  \node[font=\tiny, anchor=north east, inner sep=1pt] at (7.95,-5.68) {edge device farm};
  \node[tiny] (rstar) at (6.8,-4.53) {selected sub-network $r^{*}$ per device};
  \draw[->] (sel.south) -- (rstar.north);
  \draw[->] (rstar) -- (dj.north);
  \draw[->] (rstar) -- (dr.north);
  \draw[->] (rstar) -- (du.north);
  \draw[band] (-0.05,-5.95) rectangle (8.15,-3.25);
  \node[blab] at (-0.05,-5.95) {C. SELECT PER DEVICE};
  \end{tikzpicture}}
  \caption{Train once, adapt every candidate, select per device. A
  weight-shared supernetwork yields an 11-candidate calibrated grid;
  \emph{every} candidate is adapted to the machine manual by
  retrieval-grounded distillation. The deployed rank is selected per
  device from judged quality and anchor-measured throughput under a
  capability floor and memory budget (Eq.~\ref{eq:select}).}
  \label{fig:pipeline}
\end{figure}

The proposed pipeline (Fig.~\ref{fig:pipeline}) compresses an
instruction-tuned LLM in two distillation stages. First, the base model
becomes a weight-shared supernetwork finetuned with a sandwich-style
in-place distillation rule, so that many candidate sub-networks are
trained jointly (Sec.~\ref{sec:sandwich}). A compact sub-network is then extracted under
a deployment budget (Sec.~\ref{sec:extract}), and a hardware-grounded
selection stage decides \emph{which} candidate to commit per deployment
target (Sec.~\ref{sec:select}). Second, the extracted model is
specialized to the target application, with the unpruned teacher
distilling retrieval-grounded answers over the factory document into
the sub-network through low-rank adapters (Sec.~\ref{sec:postkd}).
The result is exported to a quantized format (8-bit ONNX for GPU
targets, 4-bit GGUF for Arm CPUs) and serves as
the reasoning core of the deployed agent (Sec.~\ref{sec:agent}).

\subsection{Supernetwork and search space}
\label{sec:space}
Supernetworks are built with an identical recipe at two scales,
Llama-3.2-3B-Instruct and Llama-3.2-1B-Instruct \cite{llama3},
matching the deployment tiers of Sec.~\ref{sec:platforms}; the 3B
instance is described here. The transformer layers are made
\emph{elastic} along two dimensions, network depth and per-layer
MLP intermediate size, which leave the shared residual
dimensionality and attention geometry intact (the costliest to
recover after pruning \cite{llmpruner,shearedllama}) and trade
quality against size smoothly. Calibration collapses this space into an 11-point
candidate grid spanning 2.20-3.21 billion unique (as-deployed)
parameters on the 3B base and 0.90-1.24 billion on the 1B; the
largest candidate is the unpruned base model itself.
A candidate is activated \emph{in place} by slicing the shared
weight tensors, so no separate student model is instantiated.

\subsection{Sandwich-style in-place distillation}
\label{sec:sandwich}
The supernetwork is finetuned on general instruction data
(Alpaca-GPT4 \cite{alpacagpt4}) before any sub-network is committed
to.
Following slimmable networks \cite{slimmable} and single-stage
supernets \cite{bignas}, each optimization step trains the
\emph{full} supernetwork together with $M{=}3$ sampled sub-networks,
using the full network's detached output distribution as an
in-place teacher. Sampling is
\emph{importance-calibrated}, in that channels are pre-permuted by
importance and the $M$ sub-networks are drawn uniformly from the
11-point calibrated grid of Sec.~\ref{sec:extract} rather than the
raw space. The
per-step objective is
\begin{equation}
\mathcal{L}_{\text{step}} =
   \mathcal{L}_{\mathrm{CE}}(z_{\text{full}}, y)
   + \sum_{i=1}^{M}
   \big[\alpha\mathcal{L}_{\mathrm{CE}}(z_{i}, y)
      + \beta\tau^{2}
        \mathrm{KL}\!\big(p^{\tau}_{\text{full}}\,\|\,p^{\tau}_{i}\big)\big],
\label{eq:sandwich}
\end{equation}
where $z_{\text{full}}$ and $z_{i}$ are the logits of the full
network and the $i$-th sub-network on the same batch, $y$ are the
ground-truth tokens, $p^{\tau}$ is the softmax softened at
temperature $\tau$, $\mathrm{KL}$ is the forward Kullback-Leibler (KL)
divergence from the full network's detached distribution, and
$\alpha,\beta$ balance the terms ($\alpha{=}0.8$, $\beta{=}0.2$,
$\tau{=}1.0$ in production). Gradients from all forwards accumulate into the
shared weights, transferring large-capacity behavior to smaller
candidates without a separate teacher pass. The ablation in
Sec.~\ref{sec:results} isolates what calibrated sampling
contributes.

\subsection{Sub-network extraction}
\label{sec:extract}
After supernetwork training, sub-networks are materialized as
standalone checkpoints. Candidates come from an 11-point grid over the
elastic dimensions, built by binning the parameter range and
keeping the best WikiText-perplexity configuration per bin plus the
smallest and the full one.
Extraction itself is adopted from the sub-network
selection literature \cite{subnetdistill}.

\subsection{Hardware-grounded sub-network selection}
\label{sec:select}
Committing to one grid rank per target is itself a selection
problem, and cheap proxies are insufficient for it. The
natural quality proxy (KL divergence to the full-width supernetwork
on production-shaped prompts) is perfectly rank-correlated with
parameter count (Spearman $\rho{=}{-}1.0$); no monotone proxy can
express the \emph{inversions} of measured fronts. Selection rests on
measurement, in three stages: (i)~proxy metrics for orientation,
(ii)~\emph{real evaluation}, in which every rank is adapted
(Sec.~\ref{sec:postkd}), exported, and measured for routing
accuracy, judged RAG quality, and on-device throughput, latency,
memory, and energy (Sec.~\ref{sec:platforms}), and
(iii)~\emph{selection}, in which the deployed rank is
\begin{equation}
r^{*} = \arg\min_{r \in \mathcal{F}}
  \big[\, w\,(1-\tilde{Q}_r) + (1-w)\,(1-\tilde{T}_r) \,\big],
\label{eq:select}
\end{equation}
where $Q_r$ is judged RAG quality, $T_r$ measured on-device
throughput, tildes denote min-max normalization over the Pareto
front, $w{=}0.5$, ties resolve toward higher throughput, and
$\mathcal{F}=\{r : C_r \ge \phi\, C_{\text{full}},\;
M_r \le M_{\max}\}$ imposes an ARC-Easy capability floor
($\phi{=}0.8$) and a memory ceiling. The floor
$\phi$ is a deployment-policy parameter, yet it changes the
outcome, disqualifying candidates retaining only $67$-$78\%$ of
the unpruned ARC-Easy score, among them the unconstrained blend's
picks on both grids. A three-anchor
throughput predictor ($\text{tok/s}=a/n_{\text{params}}$), fitted
per device on three ranks spanning the parameter range, predicts the
eight held-out ranks with $2.5$-$3.1\%$ mean absolute error, and
the fronts computed from the predicted throughput coincide with the
fully measured fronts on both grids. Commissioning a new device,
therefore, requires benchmarking only the three anchors ($6.0$ hours
instead of $28.8$ for the full grid).

\section{Retrieval-Grounded Post-Extraction Distillation}
\label{sec:postkd}

\subsubsection*{Distillation data}
The application corpus is a 187-page mill operator manual, split
into 1{,}460 overlapping 100-token chunks. Grounded question-answer pairs are first generated
per chunk with grammar-constrained JSON decoding (686 training pairs
plus a 633-question out-of-sample set). Each question is then re-answered by the \emph{teacher}, the
unpruned base instruction model at the same scale, under the
\emph{production} prompt template and retriever top-4 contexts, so
training matches inference exactly.
Following Retrieval-Aware Finetuning (RAFT) \cite{raft}, with
probability $0.5$ one or two of the four golden chunks are replaced by
\emph{distractor} chunks drawn from lower retrieval ranks, teaching
the student to ignore near-miss context.

\subsubsection*{Objective}
The extracted sub-network is adapted with low-rank adapters (LoRA)
\cite{lora} of uniform rank $r{=}8$ on attention and MLP projections,
keeping the base weights frozen. On answer tokens only, the student
minimizes
\begin{equation}
\mathcal{L}_{\text{post}} =
   \alpha'\,\mathcal{L}_{\mathrm{CE}}
   + \beta'\,\tau'^{2}\,
     \mathrm{KL}\!\big(p^{\tau'}_{T}\,\|\,p^{\tau'}_{S}\big),
\label{eq:postkd}
\end{equation}
with teacher distribution $p_{T}$, student distribution $p_{S}$, and
$\alpha'{=}\beta'{=}0.5$, $\tau'{=}2.0$ in the deployed runs at both
scales.
Tool-routing demonstrations (Sec.~\ref{sec:agent})
are mixed into the same run with the task loss only.

\subsubsection*{Consolidation and export}
After training, adapters are merged into dense weights.
The
result is exported to ONNX and reduced by post-training quantization
\cite{quantref} to blockwise INT8 weights with FP16 activations
(W8A16), the output head staying in full precision. For the Arm CPU tiers the same merged checkpoint is
instead exported to 4-bit GGUF (Q4\_K\_M) for llama.cpp, under a
verified parity contract tying the input embeddings to the output
head exactly as the ONNX build does. At these formats the 3B rank-6
occupies 1.85~GB (Q4\_K\_M) and 2.1~GB (W8A16); the 1B rank-5,
781~MB (Q4\_K\_M).

\section{Agentic Retrieval-Augmented Deployment}
\label{sec:agent}
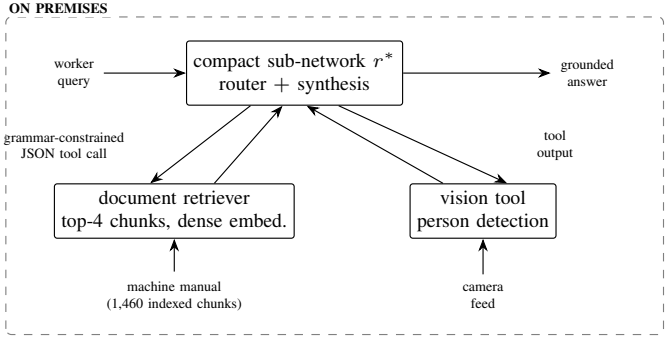
\begin{figure}[t]
  \centering
  \resizebox{\columnwidth}{!}{%
  \begin{tikzpicture}[
    font=\scriptsize, >={Stealth[length=1.6mm]},
    box/.style={draw, align=center, inner sep=2.5pt, rounded corners=1pt},
    band/.style={draw=black!50, dashed, rounded corners=2pt},
    blab/.style={font=\bfseries\tiny, anchor=south west, fill=white, inner sep=1pt},
    tiny/.style={font=\tiny, align=center},
  ]
  \node[tiny] (wq) at (0.8,0) {worker\\query};
  \node[box, minimum height=8mm] (llm) at (3.55,0)
    {compact sub-network $r^{*}$\\router $+$ synthesis};
  \node[tiny] (ans) at (7.2,0) {grounded\\answer};
  \draw[->] (wq) -- (llm);
  \draw[->] (llm) -- (ans);
  \node[box] (ret) at (2.05,-1.72) {document retriever\\top-4 chunks, dense embed.};
  \node[box] (vis) at (5.9,-1.72) {vision tool\\person detection};
  \draw[->] ($(llm.south)+(-0.55,0)$) -- ($(ret.north)+(-0.3,0)$);
  \draw[->] ($(ret.north)+(0.5,0)$) -- ($(llm.south)+(-0.15,0)$);
  \draw[->] ($(llm.south)+(0.55,0)$) -- ($(vis.north)+(0.3,0)$);
  \draw[->] ($(vis.north)+(-0.5,0)$) -- ($(llm.south)+(0.15,0)$);
  \node[tiny, anchor=east] at (1.55,-0.9) {grammar-constrained\\JSON tool call};
  \node[tiny, anchor=west] at (6.45,-0.9) {tool\\output};
  \node[tiny] (man) at (2.05,-2.78) {machine manual\\(1{,}460 indexed chunks)};
  \draw[->] (man) -- (ret);
  \node[tiny] (cam) at (5.9,-2.78) {camera\\feed};
  \draw[->] (cam) -- (vis);
  \draw[band] (-0.05,-3.26) rectangle (8.15,0.7);
  \node[blab] at (-0.05,0.7) {ON PREMISES};
  \end{tikzpicture}}
  \caption{On-premise tool-augmented RAG assistant, instantiating the
  task-orchestrator design of \cite{mfgframework}. The selected
  sub-network $r^{*}$ routes each query through a grammar-constrained
  JSON tool call and then synthesizes the grounded answer from the
  tool output.}
  \label{fig:system}
\end{figure}

The compact model is the reasoning core of an on-premise assistant
that adopts the task-orchestrator design of \cite{mfgframework}, in which the orchestrator
delegates to external agents (Fig.~\ref{fig:system}).

 Each
query is routed by the compact model itself, with
grammar-constrained decoding guaranteeing valid JSON tool calls,
either to retrieval over the machine manual (dense
embeddings \cite{bge}, top-4)
or to a vision tool built on an off-the-shelf person detector
\cite{yolo}, then synthesizes the grounded answer from the tool
output under the distillation data's prompt template
(Sec.~\ref{sec:postkd}). Models are served with
ONNX Runtime GenAI (W8A16) on GPU targets and llama.cpp (Q4\_K\_M)
on Arm CPU targets.

The full stack runs on premises (development on a single 24-GB GPU
partition), so documents and camera images never leave the site.

\section{Experimental Setup}
\textbf{Models:} the supernetworks are Llama-3.2-3B- and 1B-Instruct
(identical recipe); the deployed sub-networks are chosen per
platform by Sec.~\ref{sec:select} at the balanced quality
weight (3B rank 6 on the Jetson and the RevPi; 1B rank 5 on the UNO~Q);
the post-extraction teacher is the unpruned base at each scale.
\textbf{Data:} the common-task stage uses Alpaca-GPT4
\cite{alpacagpt4}; the application stage uses 686 manual QA pairs
with RAFT distractors and a 633-question held-out set
(Sec.~\ref{sec:postkd}), plus $n{=}40$ routing queries.
\textbf{Metrics:} RAG quality is measured by the three RAG metrics,
faithfulness, answer relevancy, and context
utilization, scored by an LLM judge (Claude Opus~4.8, medium
reasoning effort) in one interleaved single-session pass per
campaign (Sec.~\ref{sec:platforms}); efficiency by TTFT, decode TPS, peak
memory, and energy per inference and at standby; and tool routing
by accuracy. Judged scores are comparable only within a session;
absolute values differ across Tables~\ref{tab:capability},
\ref{tab:strategies} and~\ref{tab:xplatform}. The vision
tool uses an off-the-shelf detector and is reported as functional,
not benchmarked. 
\textbf{Baselines:} the unpruned base under the identical RAG
stack, the unadapted extraction, supervised finetuning, and the
KD-recipe ladder of Table~\ref{tab:stage2}.
\textbf{Hardware:} all training runs on the on-prem GPU partition of
Sec.~\ref{sec:agent}; supernetwork training dominates cost
(${\sim}51$ h 3B, $22.8$ h 1B); each Stage-2 variant of
Table~\ref{tab:stage2} is a ${\sim}12$-minute LoRA run (the whole
ladder under one GPU-hour). Deployment is evaluated on the three
edge platforms of Sec.~\ref{sec:platforms}.
\textbf{Supplementary examples:} qualitative answers with
per-question judge scores and the vision demo, at
\url{https://enakronic.github.io/llm-assistant-supplement/}.

\section{Results and Discussion}
\label{sec:results}

\subsection{Capability vs.\ model size}
Table~\ref{tab:capability} separates what pruning costs from what
distillation recovers. Under an identical retrieval stack and a
single judge pass over all $633$ held-out questions, extraction at
the deployed rank~6 drops judged answer quality by
$13.7\%$ relative to the base; the retrieval-grounded
second stage returns the model to within $4.6\%$ of the unpruned
model's judged quality, recovering two thirds of the loss
(\textbf{RQ1}). The gap
is estimated over $633$ paired questions; its bootstrap $95\%$
confidence interval is $[-5.9\%,-3.3\%]$ of the base score.
General capability and adapted task quality also decouple. ARC-Easy
falls almost linearly with parameter count ($R^{2}{=}0.93$); judged
RAG quality is only loosely size-ordered ($R^{2}{=}0.76$), with
rank~6 matching the two largest candidates. Selection exploits this
decoupling, and the capability floor guards against it.

\subsection{Stage-2 decomposition and baselines}
Table~\ref{tab:stage2} isolates the Stage-2 ingredients ($n{=}291$,
one judge pass): context-free supervised finetuning \emph{degrades}
the extracted model ($0.654$ vs.\ $0.701$), grounded task loss
reaches $0.727$, the softened distillation term adds three points,
and RAFT distractors complete the deployed recipe at $0.765$.

\begin{table}[t]
\caption{Quality vs.\ model size at the deployed rank (grid rank~6;
$n{=}633$ held-out questions, one judge pass). Quality is the mean of
the three RAG metrics; $\Delta$ is relative to the unpruned base
model under the identical RAG stack.}
\label{tab:capability}
\centering
\setlength{\tabcolsep}{4pt}
\begin{tabular}{lcccc}
\toprule
\textbf{Model} & \textbf{Params} & \textbf{Size} & \textbf{Quality} & \textbf{$\Delta$} \\
\midrule
Base 3B + RAG (W8A16)   & 3.21B & 2.42 GB & 0.848 & ref. \\
Extracted, no Stage 2   & 2.77B & 2.1 GB  & 0.732 & $-13.7\%$ \\
\;\;+ Stage-2 KD        & 2.77B & 2.1 GB  & 0.809 & $-4.6\%$ \\
\bottomrule
\end{tabular}
\end{table}
\begin{table}[t]
\caption{Stage-2 decomposition at extraction rank 4 ($n{=}291$
held-out questions, one judge pass; a different campaign from
Table~\ref{tab:capability}, so absolute values differ). Quality is
the mean of the three RAG metrics. Rows below ``Extracted'' are
recipe ablations; ``+'' rows build cumulatively, and the last row
is the deployed recipe.}
\label{tab:stage2}
\centering
\setlength{\tabcolsep}{4pt}
\begin{tabular}{lcc}
\toprule
\textbf{Configuration} & \textbf{Quality} & \textbf{$\Delta$ vs.\ base} \\
\midrule
Base (unpruned) + RAG                 & 0.791 & ref. \\
Extracted, no adaptation              & 0.701 & $-11.4\%$ \\
\;\;supervised FT, context-free QA   & 0.654 & $-17.3\%$ \\
\;\;task loss on grounded data ($\alpha{=}0.95$) & 0.727 & $-8.1\%$ \\
\;\;+ KD ($\alpha{=}\beta{=}0.5$, $\tau{=}2$) & 0.760 & $-3.9\%$ \\
\;\;+ RAFT distractors (\textbf{ours, deployed}) & \textbf{0.765} & $\mathbf{-3.3\%}$ \\
\bottomrule
\end{tabular}
\end{table}

\subsection{Efficiency trade-off and selection outcome}
\begin{figure}[t]
  \centering
  \resizebox{\columnwidth}{!}{%
  \begin{tikzpicture}[font=\scriptsize,
    dot/.style={circle, draw=black, inner sep=0pt, minimum size=4.4pt},
    odot/.style={circle, draw=black, fill=white, inner sep=0pt, minimum size=4.4pt},
    lab/.style={font=\tiny, inner sep=1pt}]
  \draw[->] (-0.1,0) -- (7.9,0);
  \draw[->] (0,-0.1) -- (0,2.62);
  \foreach \x/\t in {1.652/2.4, 2.974/2.6, 4.296/2.8, 5.617/3.0, 6.939/3.2}
    \draw (\x,0) -- (\x,-0.07) node[below, font=\tiny] {\t};
  \foreach \y/\t in {0.302/0.4, 0.907/0.5, 1.512/0.6, 2.116/0.7}
    \draw (0,\y) -- (-0.07,\y) node[left, font=\tiny] {\t};
  \node[lab] at (3.9,-0.5) {unique parameters (billions)};
  \node[lab, rotate=90] at (-0.72,1.4) {score};
  \draw[black!40] (0.330,1.131) -- (0.859,1.342) -- (1.652,0.889) -- (2.577,1.506)
    -- (3.304,1.983) -- (4.098,2.274) -- (4.891,1.911) -- (5.420,1.887)
    -- (6.147,2.207) -- (6.808,2.280) -- (7.006,2.262);
  \draw[black!40] (0.330,0.224) -- (0.859,0.242) -- (1.652,0.320) -- (2.577,0.538)
    -- (3.304,0.375) -- (4.098,0.713) -- (4.891,0.713) -- (5.420,1.070)
    -- (6.147,1.179) -- (6.808,1.318) -- (7.006,1.276);
  \foreach \p in {(0.330,1.131),(0.859,1.342),(1.652,0.889),(2.577,1.506),(3.304,1.983),(4.098,2.274),(4.891,1.911),(5.420,1.887),(6.147,2.207),(6.808,2.280),(7.006,2.262)}
    \node[dot, fill=black] at \p {};
  \foreach \p in {(0.330,0.224),(0.859,0.242),(1.652,0.320),(2.577,0.538),(3.304,0.375),(4.098,0.713),(4.891,0.713),(5.420,1.070),(6.147,1.179),(6.808,1.318),(7.006,1.276)}
    \node[odot] at \p {};
  \node[circle, draw, inner sep=0pt, minimum size=9.5pt, thick] at (4.098,2.274) {};
  \node[lab] at (4.098,2.52) {r6};
  \node[lab, anchor=west] at (0.35,1.85) {\textbullet\ judged RAG quality ($R^{2}{=}0.76$)};
  \node[lab, anchor=west] at (0.35,0.62) {$\circ$\ ARC-Easy ($R^{2}{=}0.93$)};
  \end{tikzpicture}}\\[3pt]
  \resizebox{\columnwidth}{!}{%
  \begin{tikzpicture}[font=\scriptsize,
    dot/.style={circle, draw=black, inner sep=0pt, minimum size=4.4pt},
    lab/.style={font=\tiny, inner sep=1pt}]
  \draw[->] (-0.1,0) -- (7.9,0);
  \draw[->] (0,-0.1) -- (0,2.85);
  \foreach \x/\t in {0.475/9, 2.850/10, 5.225/11, 7.600/12}
    \draw (\x,0) -- (\x,-0.07) node[below, font=\tiny] {\t};
  \foreach \y/\t in {0.185/0.5, 1.114/0.6, 2.042/0.7}
    \draw (0,\y) -- (-0.07,\y) node[left, font=\tiny] {\t};
  \node[lab] at (3.9,-0.5) {measured decode throughput (tok/s)};
  \node[lab, rotate=90] at (-0.72,1.35) {judged RAG quality};
  \draw[black!40, thick] (0.333,2.295) -- (2.518,2.281) -- (3.444,1.837)
    -- (4.228,1.109) -- (6.436,0.856) -- (7.101,0.527);
  \node[dot, fill=black]    at (1.330,2.267) {}; \node[lab] at (1.330,2.44) {1};
  \node[dot, fill=black]    at (0.333,2.295) {}; \node[lab] at (0.14,2.42) {2};
  \node[dot, fill=black]    at (0.974,2.181) {}; \node[lab] at (0.90,2.00) {3};
  \node[dot, fill=black]    at (1.473,1.692) {}; \node[lab] at (1.60,1.53) {4};
  \node[dot, fill=black]    at (2.090,1.728) {}; \node[lab] at (2.26,1.87) {5};
  \node[dot, fill=black]    at (2.518,2.281) {};
  \node[circle, draw, inner sep=0pt, minimum size=9.5pt, thick] at (2.518,2.281) {};
  \node[lab] at (3.02,2.46) {$r^{*}{=}6$};
  \node[dot, fill=black!25] at (3.444,1.837) {}; \node[lab] at (3.62,1.98) {7};
  \node[dot, fill=black!25] at (4.228,1.109) {}; \node[lab] at (4.46,1.21) {8};
  \node[dot, fill=black!25] at (4.916,0.155) {}; \node[lab] at (5.15,0.23) {9};
  \node[dot, fill=black!25] at (6.436,0.856) {}; \node[lab] at (6.44,1.03) {10};
  \node[dot, fill=black!25] at (7.101,0.527) {}; \node[lab] at (7.34,0.64) {11};
  \node[dot, fill=black]    at (0.3,0.80) {};
  \node[lab, anchor=west] at (0.45,0.80) {meets the capability floor};
  \node[dot, fill=black!25] at (0.3,0.52) {};
  \node[lab, anchor=west] at (0.45,0.52) {below the floor (disqualified)};
  \draw[black!40, thick] (0.16,0.24) -- (0.46,0.24);
  \node[lab, anchor=west] at (0.5,0.24) {measured Pareto front};
  \end{tikzpicture}}
  \caption{Top: judged RAG quality vs.\ ARC-Easy across the 3B
  grid; size predicts general capability but not adapted RAG
  utility. Bottom: the measured quality-throughput plane on the
  Jetson; gray candidates fall below the capability floor, the ring
  marks the selected rank.}
  \label{fig:pareto}
\end{figure}
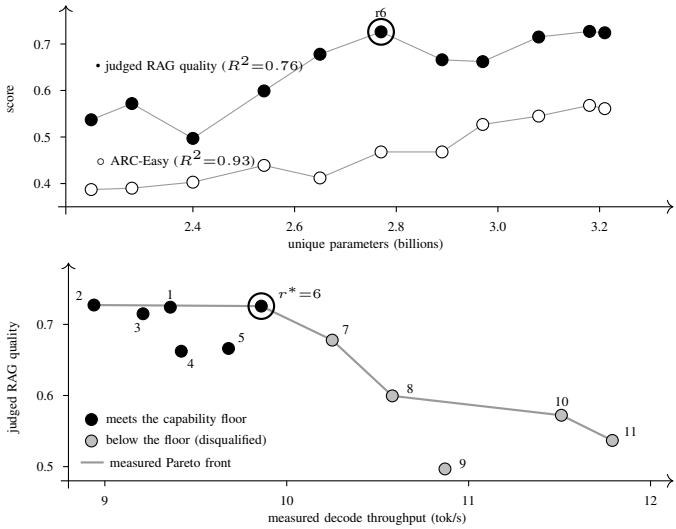

Fig.~\ref{fig:pareto} shows the decoupling and the measured
selection plane; Table~\ref{tab:strategies} compares selection
rules on the RevPi grid. Every simpler rule forfeits something:
size- and speed-driven picks retain only $69$-$74\%$ of base
capability and quality-only ignores throughput, while the
constrained blend keeps all three (\textbf{RQ2}). On the 1B grid the floor likewise removes the
unconstrained pick (rank~10) and selects rank~5.
The floor sweep (Table~\ref{tab:strategies}, bottom) shows the
picks move only at extreme values. With the floor in place, the blend weight barely
matters, selecting rank~6 at every $w\in[0.15,0.85]$ on both 3B
devices and rank~5 on the UNO~Q (rank~8 at speed-leaning $w$).
\begin{table}[t]
\caption{Top: selection strategies on the RevPi 3B grid (quality
and TPS from the all-rank judge session at the 256-token selection
cell, hence the TPS difference from
Table~\ref{tab:xplatform}'s 128-token cell; ARC ret.\ is relative
to the unpruned model, which rank~2 marginally exceeds). Bottom:
selected rank vs.\ floor $\phi$ at the balanced weight (deployed
setting in bold).}
\label{tab:strategies}
\centering
\setlength{\tabcolsep}{3.5pt}
\begin{tabular}{lcccc}
\toprule
\textbf{Strategy} & \textbf{Rank} & \textbf{Quality} & \textbf{TPS} & \textbf{ARC ret.} \\
\midrule
Smallest / fastest feasible & 11 & 0.537 & 5.23 & 69\% \\
Largest feasible (params)   & 1  & 0.724 & 3.25 & 100\% \\
Quality-only                & 2  & 0.727 & 2.99 & 101\% \\
Q/T blend, no floor         & 7  & 0.678 & 4.28 & 74\% \\
\textbf{Q/T blend + floor (ours)} & \textbf{6} & \textbf{0.726} & \textbf{3.37} & \textbf{83\%} \\
\bottomrule
\end{tabular}
\par\vspace{5pt}
\setlength{\tabcolsep}{5pt}
\begin{tabular}{lccccc}
\toprule
$\phi$ & 0.70 & 0.75 & \textbf{0.80} & 0.85 & 0.90 \\
\midrule
RevPi (3B)  & 7 & 6 & \textbf{6} & 1 & 1 \\
Jetson (3B) & 6 & 6 & \textbf{6} & 1 & 1 \\
UNO~Q (1B)  & 5 & 5 & \textbf{5} & 5 & 2 \\
\bottomrule
\end{tabular}
\end{table}

\begin{table*}[t]
\caption{Cross-platform evaluation of the deployed pipeline
(reference cell 128-token prompt / 128 generated, medians of 8).
Italic rows: percent change vs. the unpruned rank-1 model 
measured identically on the same device (same-session deltas; E2E
derived identically). Quality: $n{=}633$ (3B, bootstrap $95\%$
CI ${\pm}0.012$-$0.017$), $n{=}38$ (1B, ${\pm}0.05$-$0.08$).}
\label{tab:xplatform}
\centering
\setlength{\tabcolsep}{4pt}
\begin{tabular}{lccccccccccc}
\toprule
\textbf{Platform} & \textbf{Sub-net} & \textbf{Params} & \textbf{Faithf.} &
\textbf{Ans.\ rel.} & \textbf{Ctx.\ util.} & \textbf{TTFT (s)} &
\textbf{TPS} & \textbf{E2E (s)} & \textbf{Mem (GB)} &
\textbf{E/inf.\ (J)} & \textbf{Standby (W)} \\
\midrule
Jetson Orin Nano 8~GB & 3B r6  & 2.77B & 0.773 & 0.845 & 0.811 & 0.23 & 9.95  & 20.33  & 5.9  & 117 & 4.96 \\
\multicolumn{2}{r}{\emph{$\Delta$ vs.\ unpruned (\%)}} & $-13.7$ & $-5.5$ & $-3.9$ & $-4.3$ & $-13.2$ & $+5.3$ & $-8.6$ & $-13.0$ & $-7.8$ & $-1.2$ \\
\addlinespace[2pt]
RevPi Connect 5 4~GB  & 3B r6 & 2.77B & 0.773 & 0.845 & 0.811 & 5.63 & 4.00  & 76.87 & 2.3  & 170 & 2.48 \\
\multicolumn{2}{r}{\emph{$\Delta$ vs.\ unpruned (\%)}} & $-13.7$ & $-5.5$ & $-3.9$ & $-4.3$ & $-24.7$ & $+7.4$ & $-29.9$ & $-9.8$ & $-8.7$ & $-4.3$ \\
\addlinespace[2pt]
Arduino UNO Q 2~GB    & 1B r5  & 1.12B & 0.591 & 0.749 & 0.682 & 15.1 & 4.53  & 138.58 & 0.63 & 149 & 1.30 \\
\multicolumn{2}{r}{\emph{$\Delta$ vs.\ unpruned (\%)}} & $-9.7$ & $+6.9$ & $+1.8$ & $+5.3$ & $-11.8$ & $+13.2$ & $-11.7$ & $+2.0$ & $-7.7$ & $+4.8$ \\
\bottomrule
\end{tabular}
\end{table*}

\subsection{Grounded question answering and tool routing}
The deployed cores' scores appear in Table~\ref{tab:xplatform}. The
Jetson and RevPi cores (rank~6) reach $0.773$ faithfulness, $0.845$ answer
relevancy, and $0.811$ context utilization ($n{=}633$, the session
of Table~\ref{tab:capability}); the 1B core, judged in its own
session, scores $0.591/0.749/0.682$: the native 1B trails the compressed 3B
on all three metrics, so the 3B grid is used wherever memory allows
(the grids were judged separately, so the gap is indicative). Tool routing reaches $40/40$ 
for every fine-tuned core once $37$
routing demonstrations join the training mix, costing $1.3$-$2.5$
points across the RAG metrics.

\subsection{Ablations}
We ablate the supernetwork stage along both axes (calibrated
sampling, supernetwork training), judging all arms in one
interleaved session. At full width the arms are indistinguishable
($0.77$-$0.80$); mid-grid, where deployment candidates live,
calibrated sampling without supernetwork training collapses to
$0.44$-$0.50$ (ranks 4-5) while the trained supernetwork holds
${\sim}0.72$; supernetwork distillation, not sampling calibration,
carries mid-grid quality.
With training, the recipes split only at the extremes: calibrated
sampling is better at the largest ranks, including the deployed
rank~6 ($0.783$ vs.\ $0.738$, paired $t$-test, $n{=}30$ per rank,
$t{=}2.3$), and worse below
rank~8 ($+0.13$-$0.15$ for random sampling, $t{=}3.9$-$5.3$);
ARC-Easy reproduces the crossover. Calibrated sampling buys quality
where selection operates, at the cost of a small-rank tail it never
deploys.

\subsection{Discussion}
Retrieval is not evaluated: the retriever is fixed everywhere, so
every RAG score is conditioned on its unmeasured recall. Judged quality comes from a single LLM judge; interleaving
removes between-arm drift but not the judge itself, and the 3B and
1B grids were judged separately, so cross-grid comparisons are
indicative only. The $80\%$ ARC-Easy capability floor is defensible
but arbitrary; a different floor can change the picks, though on
our grids they move only at extreme values. Finally, the compact cores
inherit small-model failure modes. Multi-step reasoning, long tool
chains, and counting-style vision queries remain unreliable, and a
retrieval miss cannot be repaired downstream.

\section{Cross-Platform Deployment Study}
\label{sec:platforms}
Committing to a deployment target is cheap under the weight-shared
approach, since candidates are re-scored, or the recipe
re-instantiated at a smaller scale, without retraining shared
weights. We demonstrate
this by porting the complete assistant to three tiers of edge
hardware, choosing each tier's core by re-running
Sec.~\ref{sec:select}'s selection on anchor measurements from the
device itself.

\subsection{Platforms and per-target extraction}
\label{sec:platA}
\begin{figure}[t]
  \centering
  \includegraphics[width=0.40\columnwidth]{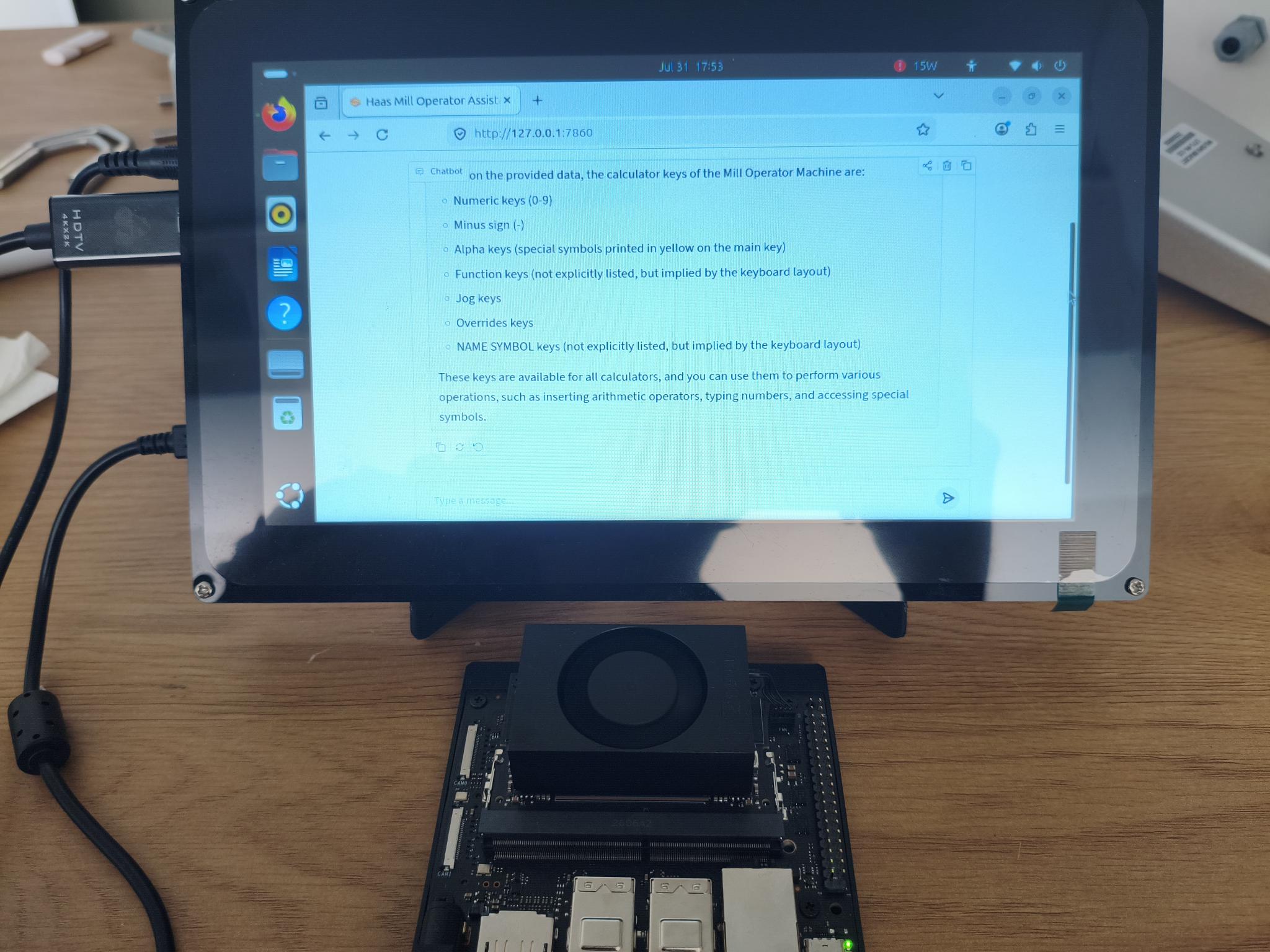}\hfill
  \includegraphics[width=0.40\columnwidth]{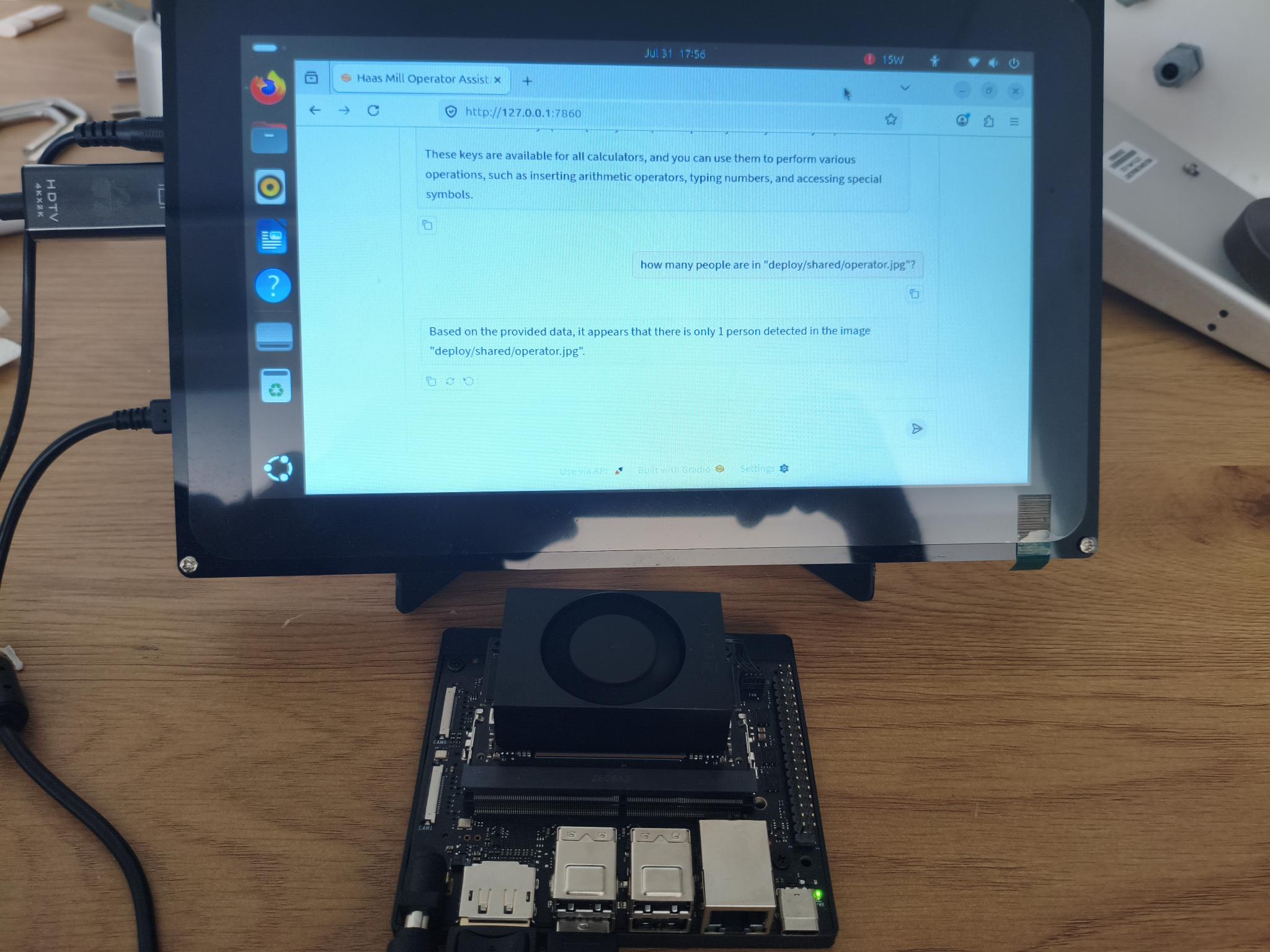}
  \caption{Demo setup on the Jetson tier. The assistant's on-device
  web interface answers a manual question through the retrieval tool
  (left) and a person-count query through the vision tool (right).}
  \label{fig:demo}
\end{figure}

We deploy on a \textbf{RevPi Connect~5} (4~GB industrial DIN-rail
computer on the Raspberry Pi CM5; llama.cpp, Q4\_K\_M, 3B rank~6),
an \textbf{NVIDIA Jetson Orin Nano} (8~GB unified memory; ONNX
Runtime GenAI, W8A16, 3B rank~6; measured in the default 15-W
mode), and an \textbf{Arduino UNO~Q} (2~GB; llama.cpp, Q4\_K\_M),
whose memory ceiling admits no 3B candidate even at 4-bit, so its
core is the 1B grid's rank~5, selected by the same procedure.
Document index, prompts, routing grammar, and queries are identical
across platforms; only model scale, quantization, and runtime
differ. Fig.~\ref{fig:demo} shows the
deployed assistant answering through both tool paths on the Jetson
tier of the demo setup.

\subsection{Metrics}
Each platform is evaluated along three axes. \textbf{(i) RAG quality:} the three RAG metrics of
Sec.~VI, scored by a frontier-LLM judge with all systems'
answers interleaved and shuffled per question in one session,
guarding against judge drift and documented position bias
\cite{llmjudge}.
Retrieval is identical everywhere, so any cross-platform difference
in the three scores comes from the reasoning core. \textbf{(ii) Latency:}
time-to-first-token (TTFT), decode tokens per second (TPS), and
end-to-end (E2E) latency per query composed from measured per-call
latencies. \textbf{(iii) Resources:} peak memory (process RSS; unified
high-water on the Orin), \emph{per-inference} joules, and
\emph{standby} watts (60-s idle, model resident).
Latency, memory, and energy use a $4\times4$ grid of prompt and
generation lengths, eight repetitions per cell after warm-up,
end-of-sequence ignored.

\subsection{Results and discussion}
\label{sec:platC}
A full query (router, retrieval, synthesis at production lengths)
takes about $20$~s on the Jetson versus one to over two minutes on the
CPU tiers, where prefill dominates (Table~\ref{tab:xplatform});
per-inference energy is similar across the three tiers
($117$-$170$~J) while standby varies by $4\times$ ($1.3$-$5.0$~W),
so at sparse duty cycles standby dominates total energy. The delta
rows compare each deployed model against the unpruned rank on the
same device; all three tiers gain on latency, energy, and memory,
except for an insignificant 2\% memory increase on the UNO~Q
(\textbf{RQ3}).

\section{Conclusion}
Application adaptation breaks size-based quality ordering, so this
paper selects per device on measured quality and throughput under a
general-capability floor; simpler selection rules demonstrably pick
worse models. In our manufacturing-manual case study, the selected
sub-networks retain ${\sim}95\%$ of the unpruned model's judged RAG
quality, route tools without error, and serve the identical
assistant from a 2-GB Arduino UNO~Q to a Jetson Orin Nano;
commissioning a new device needs only three benchmarked anchors.
Future work extends selection beyond the calibrated grid and adds a
directly measured end-to-end trace of the full agent loop.

\section*{Use of AI Tools}
Answer quality was scored by an LLM judge (Claude Opus~4.8, medium
reasoning effort) under the interleaved protocol of
Sec.~\ref{sec:platforms}. All technical content, experiments, and
analysis are the authors' own.

\end{document}